%% file: main.tex
\documentclass[sigconf]{acmart}
\hypersetup{bookmarksdepth=-2}
\usepackage{booktabs} % For formal tables
\usepackage{listings}
\usepackage{subcaption}
\usepackage{multirow}
\usepackage{soul}

\usepackage{balance}
\usepackage[shortlabels]{enumitem}

\copyrightyear{2018}
\acmYear{2018}
\setcopyright{acmlicensed}
\acmConference[ICSE-NIER'18]{40th International Conference on Software Engineering: New Ideas and Emerging Results Track}{May 27-June 3 2018}{Gothenburg, Sweden}
\acmBooktitle{ICSE-NIER'18: 40th International Conference on Software Engineering: New Ideas and Emerging Results Track, May 27-June 3 2018, Gothenburg, Sweden}
\acmPrice{15.00}
\acmDOI{10.1145/3183399.3183427}
\acmISBN{978-1-4503-5662-6/18/05}

\usepackage{color}

\definecolor{dkgreen}{rgb}{0,0.6,0}
\definecolor{gray}{rgb}{0.5,0.5,0.5}
\definecolor{mauve}{rgb}{0.58,0,0.82}

\makeatletter
\let\origsection\section
\renewcommand\section{\@ifstar{\starsection}{\nostarsection}}

\newcommand\nostarsection[1]
{\sectionprelude\origsection{#1}\sectionpostlude}

\newcommand\starsection[1]
{\sectionprelude\origsection*{#1}\sectionpostlude}

\newcommand\sectionprelude{%
	\vspace{-1ex}
}

\newcommand\sectionpostlude{%
	\vspace{-1ex}
}
\makeatother

\newcommand{\inlinedComment}[2]%{}% EMPTY for submission
{\textcolor{#1}{\small\textbf{#2}}}

\newcommand{\lx}[1]{\inlinedComment{red}{JLX says: #1}}

\begin{document}
	\title{SAR: Learning Cross-Language API Mappings with Little Knowledge}
	
	\author{Nghi D. Q. Bui}
	\affiliation{%
		\department{School of Information Systems}
		\institution{Singapore Management University}
	}
	\email{dqnbui.2016@phdis.smu.edu.sg}
	
	\author{Yijun Yu}
	\affiliation{%
		\department{Department of Computing \& Communications}
		\institution{The Open University}
	}
	\email{y.yu@open.ac.uk}
	
	\author{Lingxiao Jiang}
	\affiliation{%
		\department{School of Information Systems}
		\institution{Singapore Management University}
	}
	\email{lxjiang@smu.edu.sg}
	
	\renewcommand{\shortauthors}{Nghi et al.}

	% The default list of authors is too long for headers}
	%\renewcommand{\shortauthors}{Nghi et al.}
	%\renewcommand{\shortauthors}{Anonymous}

	\begin{abstract}
		\input{abstract}
	\end{abstract}
	
	%
	% The code below should be generated by the tool at
	% http://dl.acm.org/ccs.cfm
	% Please copy and paste the code instead of the example below.
	%
	
	%\keywords{software maintenance, language mapping, word2vec, syntactic structure, program translation}

	\maketitle
	
	\vspace*{-10pt}
	\input{body}

	\let\oldthebibliography=\thebibliography
	\let\endoldthebibliography=\endthebibliography
	\renewenvironment{thebibliography}[1]{%
		%  \vspace{10pt}
		\begin{oldthebibliography}{#1}%
			\fontsize{7.2}{7.8}\selectfont
			\setlength{\parskip}{0ex}%
			\setlength{\itemsep}{0pt}%
		}%
		{%
		\end{oldthebibliography}%
	}

	\balance
	\bibliographystyle{ACM-Reference-Format}
	\bibliography{reference}
	
\end{document}

%% file: abstract.tex
%!TEX root=sample-sigconf.tex
To save effort, developers often translate programs from one programming language to another, instead of implementing it from scratch.
Translating application program interfaces (APIs) used in one language to functionally equivalent ones available in another language is an important aspect of program translation.
Existing approaches %including MAM, StaMiner, Api2Api, and DeepAM, 
facilitate the translation by automatically identifying the API mappings across programming languages.
However, these approaches still require large amount of {\em parallel} corpora, ranging from pairs of APIs or code fragments that are functionally equivalent, to similar code comments.

To minimize the need of parallel corpora, this paper aims at an automated approach that can map APIs across languages with much less {\em a priori} knowledge than other approaches.
% not just notion, also techniques
The approach is based on an realization of the notion of {\em domain adaption}, combined with code embedding, to better align two vector spaces.
Taking as input large sets of programs, our approach first generates numeric vector representations of the programs (including the APIs used in each language),
% by using word embedding,
and it adapts generative adversarial networks (GAN) to align the vectors in different spaces of two languages.
For a better alignment, we initialize the GAN with parameters derived from API mapping seeds that can be identified accurately with a simple automatic signature-based matching heuristic.
Then the cross-language API mappings can be identified via nearest-neighbors queries in the aligned vector spaces.
% I realized: number of seeds is not a problem; the manual efforts needed to create the seeds is the problem.
%Since the number of seeds affects performance, our aim is to use fewer seeds to achieve good enough performance.
%
We have implemented the approach ({\bf SAR}, named after three main technical components in the approach) in a prototype for mapping APIs across Java and C\# programs.
%Our evaluation on a set of 2,130,264 Java files and a set of 958,394 C\# files shows that the approach can generate more API mappings accurately than other tools and deal with evolving APIs well with much fewer parallel corpora and fewer mapping seeds.
Our evaluation on about 2 million Java files and 1 million C\# files shows that the approach can achieve 54\% and 82\% mapping accuracy in its top-1 and top-10 API mapping results with only 257 automatically identified seeds, more accurate than other approaches using the same or much more mapping seeds.
%In addition, we can identify about 400 more API mappings between the Java and C\# SDKs than other approaches.

%% file: body.tex
%!TEX root=sample-sigconf.tex
\section{Introduction}
\label{sec:intro}
\input{intro}

\section{Related Work}\label{sec:related}
\input{related}

\section{Background}
\label{sec:background}
\input{background}

\section{Our Approach}\label{sec:approach}
\input{approach}

\section{Empirical Evaluation}\label{sec:expr}
\input{expr}

\section{Threats to Validity and Limitations}
\label{sec:threats}
\input{threats}

\section{Conclusion \& Future Work}
\label{sec:conclusion}
\input{conclusion}

%\begin{acks}
%	
%\end{acks}

%% file: intro.tex
%!TEX root=sample-sigconf.tex
%Language migration 
Migrating software projects from one language to another is a common and important task in software engineering. 
To support the process, various migration tools have been proposed. 
A fundamental challenge faced by such tools is to translate the library APIs of one language to functionally equivalent counterparts of another. Often, much manual effort is required to define the mappings between the respective APIs of two languages.

Several studies have addressed this API mapping problem, such as MAM~\cite{Zhong2010}, StaMiner~\cite{DBLP:conf/kbse/NNN14}, DeepAM~\cite{Gu2017}, and Api2Api~\cite{DBLP:conf/icse/NguyenNPN17}. 
% the summary is wrong: note that DeepAM doesn't require "manual" curation; also DeepAM does sequence to sequence mapping, although each sequence can be just 1.
%However, all these approaches have a shortcoming in requiring a large amount of manually curated parallel corpora in various forms. Therefore, their applicability can be constrained by the availability of such data for different languages.
MAM~\cite{Zhong2010} and StaMiner~\cite{DBLP:conf/kbse/NNN14} require as input a large body of parallel program corpora, which contain functionally equivalent code that use APIs in both languages, in order to mine the mappings.
Thus, they rely heavily on the availability of bilingual projects that implement the same functionality in two or more languages, which is not easy to find for any pair of languages. Although they rely on similar function names to reduce manual effort needed to identify parallel data, many functions with similar names may be actually functionally different, degrading the quality of training data and final mapping results.
DeepAM \cite{Gu2017} maps API sequences to sequences based on the text descriptions for the sequences. Its intuition is that two API sequences across languages may be mapped to each other if their text descriptions are similar. This approach does not need API mapping seeds, but requires many similar text descriptions across programs written in different programming languages whose availability can affect the mapping results.
%, which may not be easy to collect either.
%
Api2Api~\cite{DBLP:conf/icse/NguyenNPN17} uses a vector space transformation method inspired by Mikolov et al.~\cite{mikolov2013exploiting}, but it still requires many API mapping seeds from an external source (Java2CSharp \cite{Java2CSharp})) to map APIs across languages.
%

\begin{comment}
\lx{suggest to remove the table, even though the intention of the table is very good!
	the reason to remove: (1) some of the "seeds" may not be seeds, like DeepAM;
	(2) the number of "seed" may not be right, depending on how you count; they may not need so many to work if they focus on a particular project. 
}
\lx{refer to Table 1, say something;}
\begin{table}[b]
	\fontsize{11}{60}
	\centering
		\caption{Number of seeds used in different approaches.}
	\label{tab:pararallel_data_comparison}
	
	\begin{tabular}{|c|c|c|}\hline
		Approach                 & Num. seeds & Seed Type\\\hline\hline\hline
		MAM & $\sim$33000 & Aligned function body \\\hline
		StaMiner & $\sim$33000 & Aligned function body \\\hline
		DeepAM & $\sim$ 9 millions & <API sequence, description> pairs  \\\hline
		Api2Api & 860 & Aligned API \\\hline
		Ours & 257 & Aligned API \\\hline
	\end{tabular}
\end{table}
\end{comment}

In this paper, we propose an approach that can map APIs across languages while alleviating the shortcoming of existing approaches.
We realize that the underlying goal of state-of-the-art techniques is essentially to find a transformation that can align two different domains (in our context, the two vector spaces for APIs in two different languages).
Api2Api~\cite{DBLP:conf/icse/NguyenNPN17} is also an instance of this idea to learn an optimal transformation matrix between two vector spaces while requiring much parallel training data.
However, empirical evidence of existing approaches suggest that collecting the training data is an expensive process that requires either availability of manual inspection or high-quality documentations. This has led to the following research question we aim to answer in this paper:
"\textit{Can a model be built to minimize the need of parallel data to map APIs across languages?}".

We realize that the API mapping problem may be addressed by techniques based on generative adversarial training ~\cite{GoodfellowPMXWOCB14} with the assistance of a pre-trained model.
%\yu{Can you explain the rationale more explicitly?}
%where a complex model can be built based on a basic one that does not require much knowledge.
Given large code bases in two languages, it is likely that certain similarities between the code bases can be exploited to discover APIs of similar functionality across languages, without manually specifying parallel corpora.
Such knowledge of similar functionalities may not be big enough for a complete mapping model,
%but it can be used to {\em generate} larger number of mappings. With a high threshold of similarity, the initial mappings are almost certainly correct, and are small enough to afford human inspections. 
but it is small enough to afford human validation. 
Once validated, the knowledge can be {\em transferred} through {\em adversarial training} techniques to maximize the alignment between the two languages which results in better API mappings.

\begin{comment}
Therefore, our approach for API mapping works in following steps: 
\lx{likely too detailed for Intro; may simplify (2)(3)(4) into one, as long as mention S, A, R so that we can introduce the name SAR :-), 
	and should still mention the final near-neighbor queries for outputting API mappings.}
(1) it takes in a large number of programs in two languages, and generates a vector space representing code in each language via a word embedding technique adapted from previous studies~\cite{Zhong2010,DBLP:conf/kbse/NNN14,DBLP:conf/icse/NguyenNPN17,Gu2017,NghiICSE2018};
(2) it takes in a small number of API mapping seeds to initialize a transformation matrix between the vectors for the mapped APIs; the seeds can be automatically identified by a simple method signature-based heuristic with little manual effort;
(3) it takes in the initial transformation matrix and automatically optimizes it to maximize the alignment between all the vectors in the two vector spaces of two languages via an adversarial training technique~\cite{GoodfellowPMXWOCB14,DBLP:conf/icml/GaninL15,FacebookParallelCorpora}; and
(4) it uses frequently occurring nearest neighbors of each API to refine the transformation matrix, and this refinement step takes several iterations and validates over a training dataset until a convergence criterion is satisfied (i.e., the validation result no longer gets better).
\end{comment}

Our approach for API mapping works in the following way: 
(1) it takes in a large number of programs in two languages, and generates a vector space representing code and APIs in each language via a word embedding technique adapted from previous studies~\cite{Zhong2010,DBLP:conf/kbse/NNN14,DBLP:conf/icse/NguyenNPN17,Gu2017,NghiICSE2018};
(2) it adapts domain adaption techniques~\cite{GoodfellowPMXWOCB14,DBLP:conf/icml/GaninL15,FacebookParallelCorpora} to transform and align the two vector spaces for the two languages, with mainly three technical components: Seeding, Adversarial training, and Refinement; and
(3) it utilizes nearest-neighbors queries in the aligned vector spaces to identify the mapping result of each API. We name our approach {\bf SAR}, after the three main technical components in the domain adaption step.

We have implemented the approach in a prototype tailored for Java and C\#, and evaluated and compared it with the state-of-the-art techniques, such as StaMiner, DeepAM and Api2Api~\cite{DBLP:conf/icse/NguyenNPN17}.
We have evaluated the prototype on a dataset of more than 14,800 Java projects containing approximately 2.1 million files and 7,800 C\# projects containing approximately 958,000 files.
Our evaluation results indicate that the approach can achieve 54\% and 82\% accuracy in its top-1 and top-10 API mapping results with only 257 automatically identified seeds, more accurate than other approaches using the same or much more mapping seeds.
In addition, we also identify about 400 more API mappings between the Java and C\# SDKs than other approaches.

The main contributions of this paper are as follows:

\noindent
\begin{itemize}[nosep,leftmargin=*]
	\item We propose SAR, a new approach based on domain adaption techniques to transform and align different vector spaces across languages with the assistance of a seeding, adversarial learning, and refinement method. To the best of our knowledge, we are the first to apply the adversarial training techniques for the API mapping task.
	\item We adapt the adversarial training techniques in a number of ways to improve its alignment of the vector spaces: (1) we use nearest-neighbor queries to identify possible mapping candidates for better alignment; (2) we use a similarity-based model selection criteria  and reduce the need of known API mappings during the training of our model; and (3) we use the Procrustes algorithm to find the exact solution of the mapping matrix.
	\item We have implemented the approach and evaluated it with a corpus containing millions of Java and C\# source files; via an extensive empirical evaluation on different components of our approach, we demonstrate its advantages against other API mapping approaches in producing more accurate mappings with much fewer seeds that can be automatically identified.
\end{itemize}

The rest of the paper is organized as follows.
Section~\ref{sec:related} discusses studies in the literature closely related to this paper;
Section~\ref{sec:background} presents the background about vector space mapping and adversarial learning;
Section~\ref{sec:approach} presents our approach in detail;
Section~\ref{sec:expr} evaluates our approach to demonstrate its effectiveness and discuss its limitations; and
Section~\ref{sec:conclusion} concludes with possible future work.

%% file: related.tex
%!TEX root=sample-sigconf.tex

This section briefly reviews related work on cross-language program translation and relevant techniques.

\paragraph{Cross-Language Program Translation}
For the problem of cross-language program translation, much work has utilized various statistical language models for tokens~\cite{Nguyen2013}, phrases~\cite{Nguyen2015,Nguyen2016,Karaivanov:2014:PST:2661136.2661148}, or APIs~\cite{Zhong2013,Nguyen2014,DBLP:conf/kbse/NNN14,Phan2017,Zhong2010}.
A few studies also used word embedding for API mapping and migration (e.g., \cite{Gu2017,Gu2016,DBLP:conf/icse/NguyenNPN17,Phan2017}), but our work does not need large number of manually specified parallel corpora or mapping seeds.
Tools for translating code among specific languages in practice (e.g., Java2CSharp \cite{Java2CSharp}) also often dependent on manually defined rules specific to the grammars of individual languages, while our approach alleviates the need of language-specific rules. %\yu{Check whether this is true.}

MAM~\cite{Zhong2010} and StaMiner~\cite{DBLP:conf/kbse/NNN14} rely on the availability of bilingual projects that implement the same functionality in two or more languages. DeepAM \cite{Gu2017} requires many similar text descriptions across programs written in different programming languages whose availability can affect the mapping results. Api2Api~\cite{DBLP:conf/icse/NguyenNPN17} requires many API mapping seeds from Java2CSharp \cite{Java2CSharp}) to map APIs across languages.
The idea of our approach is most similar to Api2Api, while we combine seed-based and unsupervised domain adaptation techniques to reduce the need of mapping seeds.

\paragraph{Relevant Techniques}
For the techniques used to represent, model, learn source code, many studies exist for building various statistical language models of code for various purposes in recent years~\cite{Allamanis2018}. When it comes to what models to use for code, there is still much room for improvement.
% such as code search, bug localization, and code completion. 
Hellendoorn et al.~\cite{hellendoorn2017fse} showed that simpler code learning models (e.g., n-gram) with caches of code locality and hierarchy may outperform complex deep neural network models.
%But this may also indicate that using code locality and structural information within deep learning may further improve code learning accuracy.
While other studies (e.g., \cite{Nguyen2014,Gu2017,Karaivanov:2014:PST:2661136.2661148}) demonstrate that more grammatical and semantic code features at various levels of abstraction can be useful for more accurate models. These studies provoke us to perform code embedding with structural information, and in future to explore more semantic information for code embedding.
%For out-of-vocabulary API modeling, we get inspirations from the area of natural language processing, and combine OOV word embedding heuristics (e.g., \cite{bojanowski2017enriching,pinter2017mimicking}) with code structural information in our approach.
However, existing studies using domain adaption techniques for API mapping and translation still require the creation of mapping seeds~\cite{Phan2017,DBLP:conf/icse/NguyenNPN17}.

To transform vector spaces, studies in NLP on sentence comparison and translation involve variants of bilateral models to align the contents~\cite{wang2017bilateral}, but they require parallel corpora in two languages. Recent progresses in domain adaption alleviate the need of parallel corpora~\cite{GoodfellowPMXWOCB14,DBLP:conf/icml/GaninL15,FacebookParallelCorpora}. In an application to image learning, domain adaptation through GAN has shown benefit to transfer the models from other dataset as pre-training models when training on smaller dataset~\cite{DBLP:conf/eccv/WangWHWGR18}, which provides the technical foundation for our work.

%% file: background.tex
%!TEX root=sample-sigconf.tex
%\subsection{Word Embeddings}
%
%\n{To finish}
%
%Figure \ref{fig:skipgram} shows the network of the Skip-gram model. The hidden layers contain N neurons, which represent the size of the embeddings. The final layers contain M neurons, which represent the size of the vocabulary. Thus we have a N x M matrix, which represent for the embedding lookup table. Each row represents for the embedding of a word in the vocabulary. If we learn by using the Skip-gram model from scratch, the rows are initialized randomly 
%
%\begin{figure}[t!]
%    \centering
%    \includegraphics[width=0.2\textwidth]{skipgram.png}
%    \caption{The skip gram model}
%    \label{fig:skipgram}
%\end{figure}

%\lx{A Stage 3 comment: The introduction to the techniques is too early and out of contexts; even if a reader understands the technical details, s/he may not understand how the techniques are related to the problem we are addressing. It'd be better to introduce the details after our approach overview.}

The goal of domain adaptation is to produce a mapping matrix as an approximation of the similarities between vectors in the two spaces. This section gives a brief overview of two methods for domain adaptation: seed-based or unsupervised. Apart from the two input vector spaces, the seed-based method also requires a set of seeds as the parallel training data to learn the matrix, while unsupervised method does not: the mapping matrix can be obtained through adversarial learning assuming that similarity exists between the distributions of vectors in the two spaces.

\subsection{Seed-based Domain Adaptation}

Given two sets of embeddings have been trained independently on monolingual data, seed-based domain adaptation is to learn a mapping using the seeds {\it s.t.} their translations are close in a shared vector space.
Such an idea has been explored for word translation in NLP~\cite{mikolov2013exploiting}, and
Api2Api \cite{DBLP:conf/icse/NguyenNPN17} adapts it to learn API mappings. 

Formally, given two vector spaces, $X$ = $\{x_{1}, \ldots, x_{n} \}$ and $Y$ = $\{ y_{1}, \ldots, y_{m}\}$, containing $n$ and $m$ embeddings for two languages $L_1$ and $L_2$, and a set $S$ of seeds of API embedding pairs $\{(x_{s_i}, y_{s_i})\}_{s_i \in \{1, |S|\}}$, we want to learn a linear mapping $W$ between the source and the target space, such that $Wx_{s_i}$ approximates $y_{s_i}$.
In theory, $W$ can be learned by solving the following objective function:
\begin{equation}
\label{eq:mapping_matrix}
W^{*} \triangleq \mathop{argmin}_{W\in M \subset \mathcal{R}^{d \times d}} ||W X_{S} - Y_{S}||
\end{equation}
where $d$ is the dimension of the embeddings; $M\subset \mathcal{R}^{d \times d}$ is the space of $d \times d$ matrices of real numbers; 
$X_S \triangleq \{x_{s_i}\} \subset X$ and $Y_S \triangleq \{y_{s_i}\} \subset Y$ contain the embeddings of the APIs in the seeds, which are matrices of size $d \times |S|$.

Instead of approximating a solution using traditional stochastic gradient descent method used in Api2Api~\cite{DBLP:conf/icse/NguyenNPN17}, there exists an analytical Procrustes problem~\cite{Procruste} solved by Xing et al.~\cite{XingWLL15NACCL}, which has a closed form solution of the mapping matrix derived from the singular value decomposition (SVD) of $YX^{T}$: 
\begin{equation}
\label{eq:procruste}
\resizebox{.4 \textwidth}{!} 
{
	$W^{*} = \mathop{argmin}_{W} ||WX_{s} - Y_{s}|| = UV^{T}, with \ U\Sigma V^{T} = SVD(Y_{s}X_{s}^{T})$
}
\end{equation}
The advantage of a closed form solution is that one can get the exact solution which is better than the approximate solution of gradient descent, and is faster in computation.

With the mapping matrix $W$, 
%one can find the mapping of a query $x$ from the source space by multiplying its vector representation with $W$, i.e., 
one can use $y_x = W x$ to map a query vector $x$. The vector $y_x$ is the mapping, or adaptation, of $x$ in the target space.

\subsection{Unsupervised Domain Adaptation} % using Adversarial Learning
\label{sec:unsupervised_domain_adaptation} 
Adversarial learning has been successfully used for domain adaptation in an unsupervised manner. In particular, the Generative Adversarial Network ~\cite{GoodfellowPMXWOCB14} achieves this goal by a model which comprises a generator and a discriminator as two inter-playing components. A generator network that aims to learn real data distribution and produce fake data to fool the other component, so-called the discriminator; the discriminator network that acts as a classifier, which aims to distinguish the generated fake data from the real data. The two components are trained in a minimax fashion and would converge when the generator has maximized its ability to generate fake data so similar to the real data that the probability for the discriminator to make a mistake would be $\frac{1}{2}$. 

Conneau et al.~\cite{FacebookParallelCorpora} use this idea as a variant for the machine translation task, which achieves significantly better results than other baselines of machine translation, which would require no parallel data to train the networks. The generator, in this case, is a mapping matrix $W$, which can simply be seen as a set of parameters that need to be learned, and the discriminator is a feed-forward neural network.  We want to find a matrix $W$ as an approximation of the mapping between the two vector spaces $X$ and $Y$. In the adversarial learning setting, we aim to optimize two parameters: one is the discriminator's parameters, denoted as $\theta_D$, the other is the mapping matrix $W$. Our goal is to find the optimal value of two sets of parameters, which results that we have two objective functions in the adversarial learning setting.

\paragraph{Discriminator objective} 
Given the mapping $W$, the discriminator (parameterized as $\theta_D$) is optimized by this objective function: 
\begin{equation}
\fontsize{6.5}{1}
L_{D} (\theta_{D}|W)= -\sum_{i=1}^nlog P_{\theta_{D}} (source =1 |Wx_{i}) - \sum_{i=1}^mlog P_{\theta_{D}} (source =0 |y_{i})
\end{equation}
where $P_{\theta_D}\big(source=1 \big|v\big)$ is the probability that a vector $v$ originates from the source embedding space (as opposed to an embedding from the target space).

\paragraph{Mapping objective} 
Given the discriminator $\theta_D$, the mapping $W$ aims to fool the discriminator's ability of predicting the original domain of an embedding by minimizing this objective function:

\begin{equation}
\fontsize{6.5}{1}
L_{W} (W|\theta_{D})= -\sum_{i=1}^nlog P_{\theta_{D}} (source=0 |Wx_{i}) - \sum_{i=1}^mlog P_{\theta_{D}} (source =1 |y_{i})
\end{equation}

\paragraph{Learning Algorithm}
\label{sec:training_algo}

The discriminator $\theta_D$ and the mapping $W$ are optimized iteratively to  minimize $L_{D}$ and $L_{W}$, respectively by following the training procedure of adversarial networks proposed by Goodfellow et al.~\cite{GoodfellowPMXWOCB14}

%In fact, the adversarial learning~\cite{FacebookParallelCorpora} can be applied directly into our task. The key assumption for this to work out is based on the similarity of distributions between the two large corpora. When the two distributions are not so similar, this approach may fail to reach its optimal performance, as we will show in our evaluation.
%To address this issue, we realize that the knowledge learned from a pretrained model can be reused as the initializer for the adversarial network. As we will show later, reusing a pretrained model has a big impact on the performance, even though the pretrained model is obtained with a minimal amount of parallel data.

%% file: approach.tex
%!TEX root=sample-sigconf.tex
%\subsection{Overview}

\begin{figure}[t]
	\centering
	\includegraphics[width=0.40\textwidth]{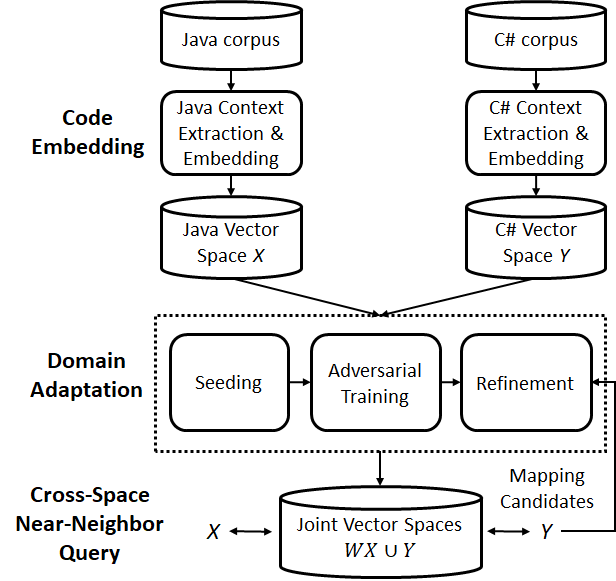}
	\caption{Approach Overview}
	\label{fig:overview}
\end{figure}

Combining the virtues of seed-based and unsupervised adversarial methods described in the background, our domain adaptation approach can approximate two spaces of vectors with minimal parallel corpora. Although unsupervised adversarial learning method does not require any seed as parallel data, the distributions of vectors (i.e. embeddings) in the two spaces may not be similar. Therefore, it is our hypothesis that the performance could be improved by initializing the unsupervised adversarial learning method with a small set of seeds taken from the seed-based domain adaptation, and by generating the rest of API mappings in the following two steps:

\begin{itemize}
	\item From large code corpora in two different languages, we create two vector spaces for APIs by adapting word embedding technique for code. From such corpora,we derive a small set of mappings based on a simple text similarity heuristic (see Code Embedding in Figure~\ref{fig:overview});
	\item The two vector spaces, along with the mapping seeds, are transformed by a mapping matrix to get aligned with each other. This step comprises three sub-steps: Seeding, Adversarial Learning, and Refinement (see Domain Adaptation in Figure~\ref{fig:overview}). % The details of each sub-step are presented in Section~\ref{sec:domain_adaptation}.
\end{itemize}

For any given API $a$ in the source language and its continuous vector representation $x$, we can map it to the other domain space by computing $y_x = W x$.
Then, one can find the top-k nearest neighbors of $y_x$ in the target vector space, using cosine similarity as the distance metric, and finally can retrieve the list of APIs in the target language that have the same embeddings as the top-k nearest neighbors. The list of APIs can then be used as the mapping results for $a$ (see Cross-Space Near-Neighbor Query in Figure \ref{fig:overview}).

The following subsections detail these sub-steps.

\begin{comment}
\begin{enumerate}[leftmargin=*]
\item \textbf{Sequence Extraction}: extracting the API sequence for training;
\item \textbf{Learning the Translation Matrix}: to learn a matrix that serves as a bridge to translate between two vector spaces;
\begin{itemize}
\item \textbf{Domain Adaptation}: to reduce the differences of the two distributions;
\item \textbf{Refinement Algorithm}: to close the gap between the unsupervised method and the unsupervised method.
\end{itemize}

\item \textbf{OOV token estimation}: To estimate the OOV APIs for software evolution
\yu{Where do you discuss OOV tokens?}
\end{enumerate}
\end{comment}

\subsection{Code Embedding via Word Embedding}\label{step:codeembedding}

We first parse source code files into Abstract Syntax Trees (AST) using {\tt srcML}~\cite{DBLP:conf/icsm/CollardDM13} for both Java and C\# projects. Then we extract from individual functions the \textit{code sequences} and perform a normalization. The normalization enriches the code sequence with structural semantic information extracted from parsing, which constitutes two steps: 

\paragraph{Filtering out noisy tokens} Tokens are considered noisy if they are not API tokens. To leverage as much structural information as possible, language keywords and AST node types are still kept for code embedding.
%Therefore, we only keep the tokens that have the rich semantic structure for the API mapping task so that the follow-up learning steps may utilize the information that is otherwise implicit in raw code.

\paragraph{Converting raw API tokens into signatures} This step reduces the variance of vocabulary existing in the source code. For example, one may extract the `List.add' method from the `java.util.List' class, or from the `com.google.common.collect.List' in an external third-party library. Even though these two APIs have the same class and method names, their usages and semantics are different. To handle such cases, we propose this additional step to convert a raw API token to its signature in qualified name format `<Package>.<Class>.<Method>'. 

Below shows an example of the normalization for the code token sequence:
%\lx{it's unclear if it's really "raw" code"}
\begin{lstlisting}
    List.add List.add if List.addAll else HashMap.put return 
==> java.util.List.add java.util.List.add if java.List.addAll 
    else java.util.HashMap.put return
\end{lstlisting}

From the corpora of code sequences, we use the skip-gram word2vec~\cite{Mikolov2013word2vec} model to train the embedding of each token. Given a large corpus as the training data, the tokens appearing in the same context would usually have their embeddings close by distance in the vector space.

\subsection{Domain Adaptation}\label{sec:domain_adaptation}

\begin{figure}[t!]
	\centering
	\includegraphics[width=0.48\textwidth]{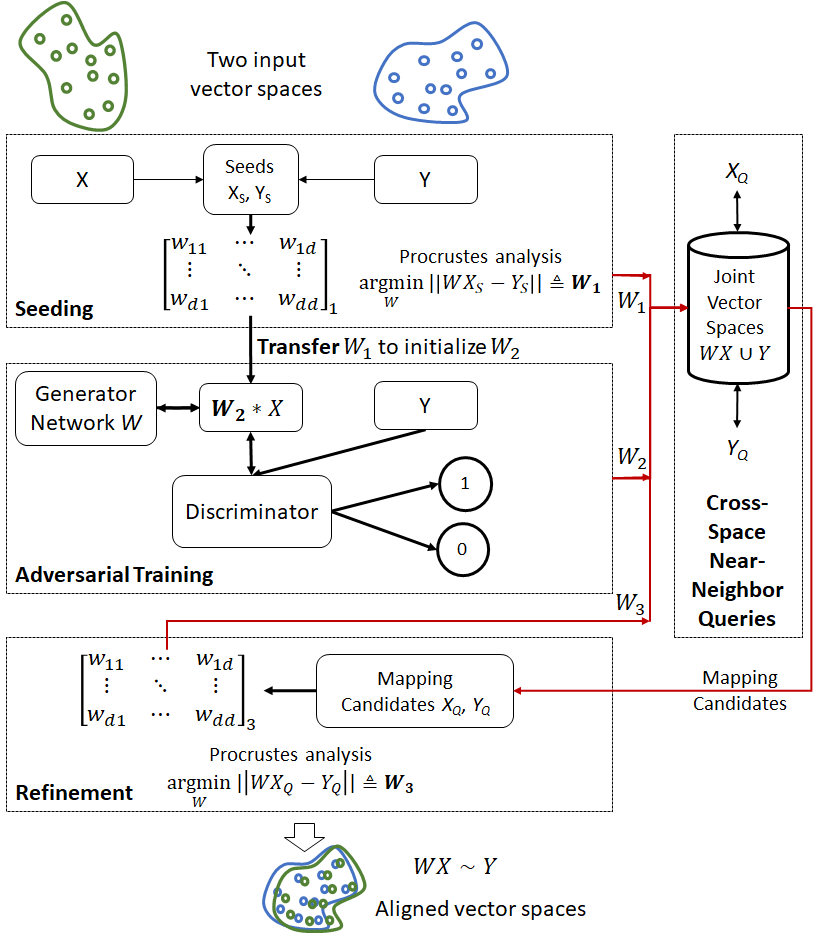}
	\caption{Domain adaptation steps to align two vector spaces}
	\label{fig:transfer_learning}
\end{figure}

%The objective of domain adaptation is to align the vector space for one language with the other so that the vector representing each API in one language is close to the vector representing the API that belongs to the other language. 
%\yu{Feels like a repeated sentence.}

Our domain adaptation comprises three steps: seeding, adversarial training, and refinement (hence the abbreviation SAR of our approach). Seeing SAR from outside as a black-box, it receives two vector spaces and a set of seeds as input and generates a mapping matrix $W$ as output. Internally, each step of SAR is a different way to improve the mapping matrix, which receives the matrix output from the previous step as input and produces the improved version of it as output. We assign $W_1$, $W_2$ and $W_3$ as the output matrix for the three steps, respectively. Figure~\ref{fig:transfer_learning} summaries the domain adaptation procedure as a whole.  The rationale for each step is described as follows: (1) \textbf{The Seeding} step to initialize a mapping matrix between the two vectors spaces based on some prior knowledge (i.e., seeds) (2) \textbf{The Adversarial Learning} step to re-use the knowledge learned from the Seeding step as an initializer for adversarial training in order to maximize the similarity between the two vector spaces (or two distributions); and (3) \textbf{The Refinement} step to make the mapping matrix better and reach its optimal state.

\subsubsection{Seeding}\label{sec:alignment}

After Code Embedding, two vector spaces are obtained to produce a mapping matrix that approximates the two vector spaces by using the knowledge from mapping seeds in a dictionary. Notice that by a simple signature-based comparison to identify APIs having the same signature name, one can identify many high-quality mapping candidates to be used as the seeds without any human effort to verify because developers often use the same name for the same functionality even when they are in different languages.

Having the dictionary $D$ obtained, in addition to the two vector spaces $X$ and $Y$, the  initial mapping matrix produced is $W_1$ by solving the Equation~\ref{eq:mapping_matrix} in Section~\ref{sec:background} (also see Seeding in Figure~\ref{fig:transfer_learning}). This seeding step can be seen as a function $A$, which will receive these three inputs and produces a transforming matrix $W_1$ such that $W_1 = A(X, Y, D)$. Internally, $A$ solves the optimization problem described in Section~\ref{sec:background} given the three inputs.

\subsubsection{Adversarial Learning}\label{sec:adversarial_learning}
The quality of the matrix $W_1$ learned in previous step is limited by the number of seeds one can provide, which results in an approximation between the source and target domains. In this case, the knowledge learned for $W_1$ can be seen as a pre-trained model and can be reused for the other model.

Formally, given the two original vector spaces, $X$ = $\{x_{1}, \ldots, x_{n} \}$ and $Y$ = $\{ y_{1}, \ldots, y_{m}\}$,containing $n$ and $m$ API embeddings obtained from the Code Embedding step, we want to find the matrix $W_2$ to \textit{maximize} the approximation of the mapping between the two vector spaces. We use the adversarial learning to achieve this goal. We build the adversarial learning network comprises of two steps: the mapping matrix $W_2$ and a discriminator network as described in Section \ref{sec:unsupervised_domain_adaptation}. Our goal is to find the optimal value of  
$W_2$ and $\theta_D$ (discriminator parameters) We achieve this by training the adversarial network with the objective functions as described in Section \ref{sec:unsupervised_domain_adaptation} to find $W_2$ and $\theta_D$.

The key difference with the general adversarial setting described Section \ref{sec:unsupervised_domain_adaptation} is that we do not initialize $W_2$ randomly as one usually does when training a neural network. Instead, we use $W_1$ as a pre-trained model to initialize for the $W_2$ so that $W_2$ is initialized with some good knowledge, even if it is small (see Adversarial Learning in Figure~\ref{fig:transfer_learning}). This step is essential to improve the performance of the API mapping results.

\paragraph{Model Selection Criteria}
To train the adversarial networks, like any other neural network architecture, we need a validation set to select the best model for the prediction step. The validation set is used to minimize overfitting when training the neural network. Concretely, for each training epoch, one needs to evaluate against the validation dataset to pick the model that has the highest validation accuracy through training. Our goal is to use as little parallel data as possible to build the model. In practice, one only has a very small number of seeds inferred from the signature-based matching, or in the worst case, one cannot infer any seed to have data for validation. As such, it is impractical to use a parallel dataset as a validation set to train neural networks in the adversarial learning step, i.e., involving additional prior knowledge.

To address this issue, we propose to use a model selection using an unsupervised criteria that quantifies the closeness of the source and target embedding spaces. Specifically, we consider them a set of K most frequent source APIs and multiply them with the mapping matrix $W$ to generate a target API mapping for each of them. We then compute the average cosine similarity between these deemed mapping and use their average as a validation metric.

\subsubsection{Refinement for Better Alignment}\label{sec:refinement_step}

%\begin{algorithm}[h]
%    \KwIn{X}
%    \KwIn{Y}
%    \KwIn{W}
%    
%    \Repeat{Converge condition}{
%        
%        
%        $C \leftarrow PRODUCE\_CANDIDATES(X,Y,W)$\
%        
%        $W \leftarrow LEARN\_MAPPING\_MATRIX(X,Y,C)$\
%        
%    }
%    \caption{{\bf Refinement Algorithm} \label{Refinement Algorithm}}
%    \label{algo:refinement}
%    \n{Not sure if this pseudocode is necessary to explain for the refinement}
%\end{algorithm}

The adversarial approach tries to align all words irrespective of their frequencies. However, rare tokens have embeddings that are less updated and are more likely to appear in different contexts in each corpus, which makes them harder
to align~\cite{FacebookParallelCorpora}. To address this problem, we use the method proposed in ~\cite{FacebookParallelCorpora} to infer a list of mapping candidates using only the most frequent tokens. Moreover, other heuristics are introduced to infer another candidate set of mapping based on the threshold of cosine similarity, which can be used as another synthetic dictionary that can combine with the top-K frequency mapping candidates.

Following the step shown in~\cite{FacebookParallelCorpora}, it is possible to build a set of mapping candidates using $W_2$ just learned with adversarial training. Assume that one can induce a combined set of mapping candidates from different heuristics above, and the quality of the combined set is good, then this set of candidates should be used to learn a better mapping and, consequently, an even better set of candidates for the next iteration. The process can repeat iteratively to obtain a hopefully better mapping and candidates set each iteration until some convergence criteria are met.  Formally, the refinement step receives $W_2$ from the previous adversarial learning step, along with the two original embeddings $X$ and $Y$ to produce the next $W_3$ iteratively (see Refinement in Figure~\ref{fig:transfer_learning}).

Specifically, we produce the mapping candidates for refinement based on two heuristics:
\begin{description}[leftmargin=*]
	\item[Top-K Frequency:] Conneau et al.~\cite{FacebookParallelCorpora} shows that by taking the top-k frequent words and their nearest neighbors in the transformed vector spaces, it can provide high-quality mapping candidates because the most frequently used words are likely to be the same across languages. Therefore, we can use the top-k frequent API names to induce the seeds for the refinement.
	\item[Cosine Similarity Threshold:]     
	Since finding API mappings in the aligned vector space is essential to finding APIs close enough in the vector space, all API pairs ``similar enough'' in the vector space aligned by Adversarial Learning can be good candidates for the refinement step.
	In this work, we use the cosine similarity as the metric to measure how similar two vectors are.
	We note that {\em not} all APIs in a language can have a mapping in another language. In the empirical case study, we show how a good threshold is found in Section~\ref{sec:refinement_study}.
\end{description}

Therefore, we can infer two sets of synthetic mapping candidates from the above heuristics. In fact, there are different ways to merge them into one single set as they can overlap as, e.g., (1) the union of the two sets, (2) the intersection of the two sets.

The matrix $W_3$ in this step is the final output of the domain adaptation process. When it comes to the step to produce the mapping from the source query, the embeddings of the query will be multiplied with $W_3$ in order to obtain corresponding mappings in the target language.

%% file: expr.tex
%\subsection{Research Questions}
We have conducted extensive empirical evaluations on our approach in various settings to answer the following research questions:
% to evaluate our approach in an ablation study:

\begin{enumerate}[RQ1,leftmargin=*]    
	\item Compared to related methods, is our approach more effective in identifying API mappings?
	\item How well do different combinations of refinement heuristics improve the performance?
	\item How do the seeds overlapping effect on the performance?
	\item What is the impact of each component in our approach on the performance?
\end{enumerate}

\subsection{Dataset}\label{sec:dataset}
We use the Java Giga corpus data described by Allamanis et al.~\cite{DBLP:conf/msr/2013}.
It involves approximately 14,807 Java projects from Github and contains approximately 2.1 millions of files. For C\#, we clone the projects on Github that have at least 1 star and collect 7,841 C\# projects with about 958,000 files.

As the main advantage of our approach, there is no need to specify which code in Java is functionally equivalent to which code in C\#. For each function in a file, we traverse the AST of the function to extract the API call sequences. For Java, we get a corpus containing 6.7 million code sequences; for C\#, we get a corpus containing 5.1 million code sequences.

%62 + 116 + 3 + 12 + 9 + 81 = 283
For evaluation, we take 860 method API mappings and 430 class API mappings defined in Java2CSharp~\cite{Java2CSharp} as the ground truth for evaluating our approach against the baselines.

\subsection{Implementation}
\label{sec:implementation}
%\subsubsection{Settings}
%\label{sec:settings}

\begin{table}
	\fontsize{7}{8}\selectfont 
	\centering
	\caption{Example of Seeds from the Signature-based Matching Heuristic}
	\label{tab:similar_class_method}
	\begin{tabular}{|c|c|}\hline
		Java                 & C\# \\\hline\hline\hline
		java.lang.\textbf{String.equals }& System.\textbf{String.Equals}  \\\hline
		java.util.\textbf{List.remove} & System.Collections.Generic.\textbf{List.Remove} \\\hline
		java.util.\textbf{Random.nextDouble} & System.\textbf{Random.NextDouble}\\\hline
		java.lang.\textbf{Math.round} & System.\textbf{Math.Round} \\\hline
		java.io.\textbf{File.Exists} & System.IO.\textbf{File.Exists} \\\hline
		
	\end{tabular}
\end{table}

We adapt {\tt Gensim}~\cite{rehurek_lrec} in NLP to produce the embeddings of tokens for the Java and C\# corpora. We use the same settings used by Mikolov et al.~\cite{mikolov2013distributed} during the training: stochastic gradient descent with a default learning rate of 0.025, negative sampling with 30 samples, skip-gram with a context window of size 10, and a sub-sampling rate of value $1e^{-4}$. 

\paragraph{Evaluation Metrics}

We define the \textit{top-k accuracy} as the evaluation metric throughout the experiments.  The top-k accuracy is defined as follow: For a test JDK API j, SAR produces a resulting list. If the true mapping API in C\# .NET for j is in the top-k resulting
list, we count it a hit. If not, we count it a miss. Top-k accuracy is computed as the ratio between the number of hits and the total of hits and misses for a given ground-truth test set.
%We use this simplified metric, as a false positive implies a false negative in this case and precision and recall would be the same with respect to the ground-truth set. 
We use this simplified metric for easier comparison with other approaches.
In real-world uses, one may retrieve a list of API mapping results given a query, and better use other information retrieval metrics, such as Mean Average Precision (MAP) or Mean Reciprocal Rank (MRR) as the evaluation metrics.

\paragraph{Code Embedding}
From the two code corpora, we scan through all pairs of APIs in the two corpora to produce a set of seeds using the signature-based matching heuristic. We got 257 seeds for this step. Table~\ref{tab:similar_class_method} shows examples of the seeds. Among these 257 seeds, we found that 83 seeds overlap in 860 mappings of the ground truth. Then, we apply the Code Embedding step on the corpora get the source embedding and target embedding, we use them, along with the seeds as the input for the domain adaptation process. 

\paragraph{Domain Adaptation}
For the seeding step, we find $W_1$ by using the Procrustes solution in Equation~\ref{eq:procruste} with the three inputs: source embedding  $X$ (Java), target embedding $Y$ (C\#) and 257 seeds. This step gives us the mapping matrix $W_1$. We implement the adversarial learning by using {\tt PyTorch}~\cite{paszke2017automatic}. We use Momentum Gradient Descent method~\cite{DBLP:conf/icml/SutskeverMDH13} to search for the optimal transformation matrix.

\begin{figure}[t!]\centering
	\vspace*{-3ex}
	\includegraphics[width=0.5\textwidth]{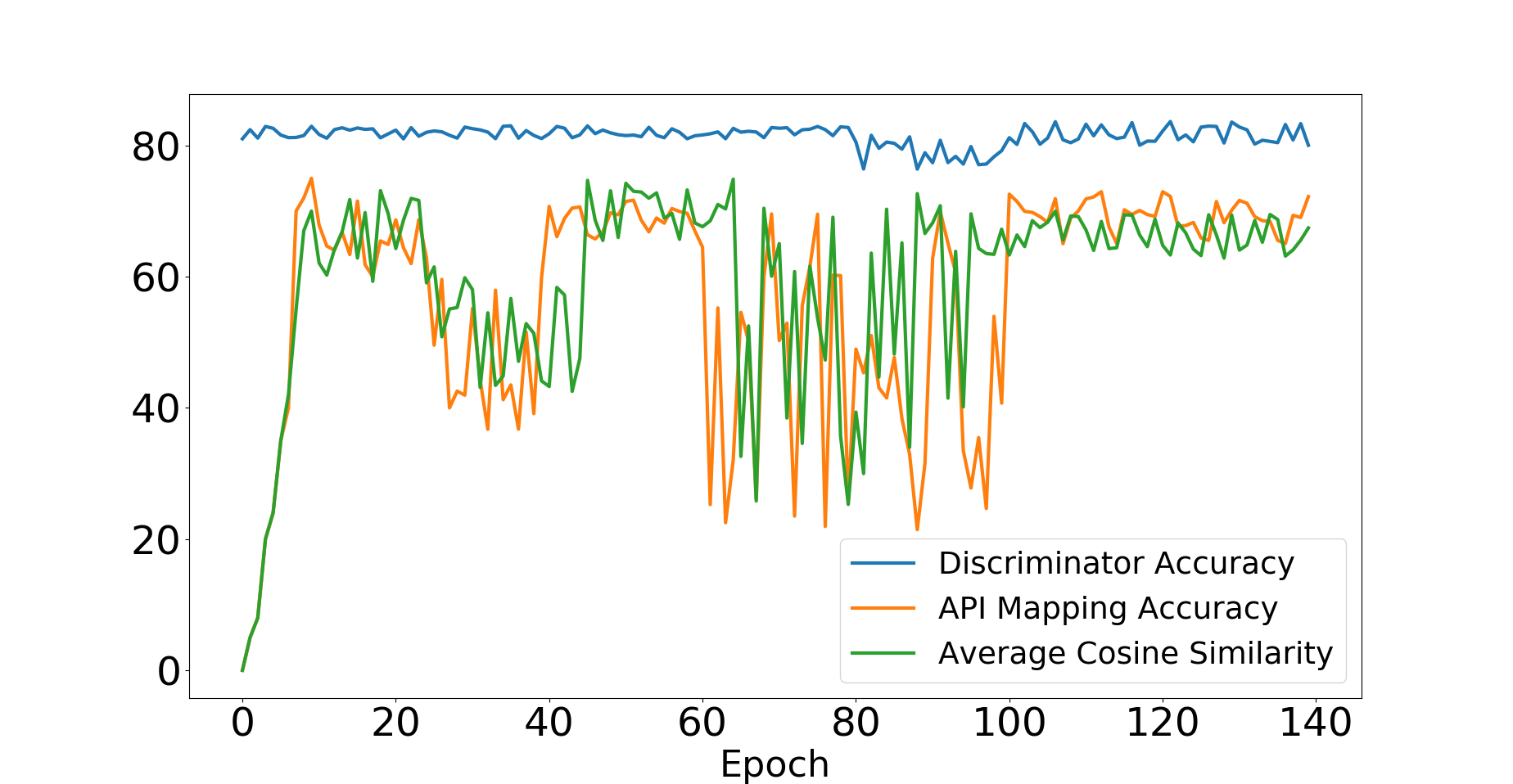}
	\caption{Unsupervised Model Selection Criteria}\label{fig:unsupervised_criterion}
\end{figure}

We use the unsupervised model selection criteria proposed in Section \ref{sec:adversarial_learning} to select the best model by choosing the top 1000 frequent API token pairs, e.g top-1 frequent token in the source  is aligned with top-1 frequent token in the target as the validation set, then we extract the $W_2$ from the model. 
Figure~\ref{fig:unsupervised_criterion} shows three different lines: (1) the discriminator accuracy, which is the accuracy in classifying the samples from the source and target embeddings (2) the API mapping accuracy, which is the accuracy when using the model to evaluate against the 1000 pairs validation set; and (3) the average cosine similarity of all the pairs. As shown, the criteria correlate well to the mapping accuracy.

From $W_2$ resulting from the adversarial training, we obtain the final $W_3$ by performing the refinement step on the basis of two heuristics in Section~\ref{sec:refinement_step}. 
For the top-N frequency heuristics, 
%by some empirical experiments,\lx{may discuss in threats to validity} 
we choose top-500 %\lx{1000? really? this is a lot, more than 860 used by api2api}
frequent tokens for the synthetic dictionary, as suggested in~\cite{FacebookParallelCorpora}.
For the second similarity threshold rule, we use 0.7 as the threshold as shown in Section~\ref{sec:refinement_study}, we found that this number balances coverage and precision of API mappings well.

Our source code and experimental results can be accessed at the anonymous repository\footnote{\url{https://github.com/djxvii/fse2019}}.

\subsection{Evaluation}

%\begin{table}
%    \centering
%    \fontsize{8}{10}\selectfont 
%    \caption{Effectiveness in mining API mappings}
%    \label{tab:approach_effectiveness}
%    \begin{tabular}{|c|c|c|c|}\hline
%        Seeds & Top-1 & Top-5 & Top-10 \\\hline\hline\hline
%        
%        
%        25  & 0.30 & 0.35 & 0.40\\ \hline
%        50  & 0.37 & 0.41 & 0.48 \\ \hline
%        100 & 0.43 & 0.50 & 0.67 \\ \hline
%        \textbf{257} & \textbf{0.54} & \textbf{0.75} & \textbf{0.82} \\\hline
%    \end{tabular}
%\end{table}

\subsubsection{RQ1. Effectiveness of SAR in Mining API Mapping}

The first question we want to answer is how effective our approach in identifying API mappings from the two vector spaces. We compare SAR with Api2Api, StaMiner, and DeepAM.

\begin{table}
	\centering
	\fontsize{6.6}{8}\selectfont 
	\caption{API Mappings - Baselines}
	\label{tab:baselines}
	\begin{tabular}{|c|c|c|c|c|c|c|}
		\hline
		Index       & Baselines                                 & K-folds          & Seeds        & Top-1         & Top-5         & Top-10        \\ \hline
		1           & \multirow{4}{*}{Random seeds: Api2Api}    & -                & 0            & 0.03          & 0.05          & 0.1           \\ \cline{1-1} \cline{3-7} 
		2           &                                           & -                & 10           & 0.09          & 0.12          & 0.14          \\ \cline{1-1} \cline{3-7} 
		3           &                                           & -                & 50           & 0.14          & 0.19          & 0.22          \\ \cline{1-1} \cline{3-7} 
		4           &                                           & -                & 100          & 0.19          & 0.24          & 0.32          \\ \hline
		5           & \multirow{4}{*}{Random seeds: SAR}        & -                & 0            & 0.25          & 0.30          & 0.35          \\ \cline{1-1} \cline{3-7} 
		6           &                                           & -                & 10           & 0.28          & 0.35          & 0.40          \\ \cline{1-1} \cline{3-7} 
		7           &                                           & -                & 50           & 0.26          & 0.43          & 0.47          \\ \cline{1-1} \cline{3-7} 
		8           &                                           & -                & 100          & 0.44          & 0.50          & 0.69          \\ \hline
		9           & \multirow{4}{*}{K-Fold: Api2Api}          & 1-fold           & 172          & 0.24          & 0.35          & 0.41          \\ \cline{1-1} \cline{3-7} 
		10          &                                           & 2-folds          & 344          & 0.34          & 0.45          & 0.55          \\ \cline{1-1} \cline{3-7} 
		11          &                                           & 3-folds          & 516          & 0.37          & 0.51          & 0.67          \\ \cline{1-1} \cline{3-7} 
		12          &                                           & 4-folds          & 688          & 0.43          & 0.64          & 0.72          \\ \hline
		13          & \multirow{4}{*}{K-Fold: SAR}              & 1-fold           & 172          & 0.36          & 0.39          & 0.48          \\ \cline{1-1} \cline{3-7} 
		14          &                                           & 2-folds          & 344          & 0.45          & 0.50          & 0.61          \\ \cline{1-1} \cline{3-7} 
		15          &                                           & 3-folds          & 516          & 0.54          & 0.66          & 0.71          \\ \cline{1-1} \cline{3-7} 
		\textbf{16} &                                           & \textbf{4-folds} & \textbf{688} & \textbf{0.59} & \textbf{0.77} & \textbf{0.84} \\ \hline
		17          & \multirow{4}{*}{Signature-based: Api2Api} & -                & 25           & 0.12          & 0.16          & 0.18          \\ \cline{1-1} \cline{3-7} 
		18          &                                           & -                & 50           & 0.20          & 0.23          & 0.29          \\ \cline{1-1} \cline{3-7} 
		19          &                                           & -                & 100          & 0.27          & 0.32          & 0.38          \\ \cline{1-1} \cline{3-7} 
		20          &                                           & -                & 257          & 0.35          & 0.41          & 0.55          \\ \hline
		21          & \multirow{4}{*}{Signature-based: SAR}     & -                & 25           & 0.30          & 0.35          & 0.40          \\ \cline{1-1} \cline{3-7} 
		22          &                                           & -                & 50           & 0.37          & 0.41          & 0.48          \\ \cline{1-1} \cline{3-7} 
		23          &                                           & -                & 100          & 0.43          & 0.50          & 0.67          \\ \cline{1-1} \cline{3-7} 
		\textbf{24} &                                           & \textbf{-}       & \textbf{257} & \textbf{0.54} & \textbf{0.75} & \textbf{0.82} \\ \hline
	\end{tabular}
\end{table}

\paragraph{Result Summary.}
Index 24 in Table~\ref{tab:baselines} uses 257 API mappings automatically selected by the signature-based matching heuristic and test against the 860 ground truth mappings.
Index 16 uses 688 mappings selected randomly from the 860 ground truth set and test against the rest.
The performance of SAR in term of top-k accuracy is shown.  As one can see in both cases, the top-1 accuracies are above 50\%, and the top-10 accuracies are above 80\%. 

\begin{table*}
	\fontsize{6.5}{7.4}\selectfont 
	\centering
	\caption{Accuracy of 1-1 Mapping when compares with StaMiner and MAM}
	\label{tab:1_1_mapping}
	\begin{tabular}{|c|c|c|c|c|c|c|c|c|c|c|c|c|c|c|c|c|c|c|}
		\hline
		\multirow{3}{*}{Package} & \multicolumn{9}{c|}{Class Mapping}                                                                            & \multicolumn{9}{c|}{Method Mapping}                                                                          \\ \cline{2-19} 
		& \multicolumn{3}{c|}{Precision}      & \multicolumn{3}{c|}{Recall}        & \multicolumn{3}{c|}{F-Score}       & \multicolumn{3}{c|}{Precision}     & \multicolumn{3}{c|}{Recall}        & \multicolumn{3}{c|}{F-score}       \\ \cline{2-19} 
		& Sta     & DeepA   & SAR             & Sta    & DeepA           & SAR     & Sta    & DeepA   & SAR             & Sta             & DeepA   & SAR    & Sta    & DeepA           & SAR     & Sta    & DeepA   & SAR             \\ \hline
		java.io                  & 70.0\%  & 80.0\%  & 80.0\%          & 63.6\% & 75.0\%          & 75.0\%  & 66.6\% & 72.7\%  & 77.5\%          & 70.0\%          & 66.7\%  & 75.0\% & 64.0\% & 87.5\%          & 82.9\%  & 66.9\% & 75.2\%  & 78.7\%          \\ \hline
		java.lang                & 82.5\%  & 80.0\%  & 85.0\%          & 76.7\% & 81.3\%          & 80.2\%  & 79.5\% & 80.7\%  & 82.6\%          & 86.7\%          & 83.3\%  & 81.5\% & 76.5\% & 87.2\%          & 78.4\%  & 81.3\% & 85.4\%  & 79.9\%          \\ \hline
		java.math                & 50.0\%  & 66.7\%  & 66.7\%          & 50.0\% & 66.7\%          & 66.7\%  & 50.0\% & 66.7\%  & 66.7\%          & 66.7\%          & 66.7\%  & 66.7\% & 66.7\% & 66.7\%          & 66.7\%  & 66.7\% & 66.7\%  & 66.7\%          \\ \hline
		java.net                 & 100.0\% & 100.0\% & 89.0\%          & 50.0\% & 100.0\%         & 100.0\% & 66.7\% & 100.0\% & 94.5\%          & 100.0\%         & 100.0\% & 100\%  & 33.3\% & 100.0\%         & 83.3\%  & 50.0\% & 100.0\% & 81.7\%          \\ \hline
		java.sql                 & 100.0\% & 100.0\% & 100.0\%         & 50.0\% & 100.0\%         & 90.0\%  & 66.7\% & 100.0\% & 95.0\%          & 100.0\%         & 50.0\%  & 75.0\% & 50.0\% & 66.7\%          & 70.0\%  & 66.7\% & 57.2\%  & 72.4\%          \\ \hline
		java.util                & 64.7\%  & 69.6\%  & 83.6\%          & 71.0\% & 72.7\%          & 75.7\%  & 67.7\% & 71.1\%  & 79.7\%          & 63.0\%          & 64.3\%  & 64.8\% & 54.8\% & 85.7\%          & 89.0\%  & 58.6\% & 73.5\%  & 76.9\%          \\ \hline
		All                      & 77.9\%  & 82.7\%  & \textbf{84.0\%} & 60.2\% & \textbf{82.6\%} & 81.3\%  & 66.2\% & 81.9\%  & \textbf{82.7\%} & \textbf{81.1\%} & 71.9\%  & 76.1\% & 57.6\% & \textbf{82.3\%} & 78.38\% & 65.0\% & 76.3\%  & \textbf{77.2\%} \\ \hline
	\end{tabular}
\end{table*}

\paragraph{Compare to Api2Api} 
The method used in Api2Api is corresponding to the seeding step in our domain adaptation process, which finds a mapping matrix by solving the Equation~\ref{eq:mapping_matrix} given a large set of seeds. We use the top-k accuracy as the evaluation metric.

Table~\ref{tab:baselines} shows the top-$k$ accuracy of our approach when comparing to Api2Api in various settings. First, we compare Api2Api with SAR using the seeds coming from two different sources: the 860 mappings defined by Java2CSharp and the 257 mappings inferred from the signature-based matching. Here we described the variances as results shown in Table~\ref{tab:baselines}, indicating that our approach can use \textit{much} fewer number of seeds compared to Api2Api but still achieve better results.
\begin{itemize}[leftmargin=*]
	\item Select randomly: we select a subset of mappings $r$ randomly from 860 mappings in the ground truth, and test against the rest $860 - r$ mappings. Concretely, $r = $ 0, 10, 50, and 100;
	\item $k$-fold: we divide the 860 mappings into $k=1,2,3,4$ folds and perform the variants of five-fold cross-validation: while $k$ folds are used as training data, the other $5-k$ folds are used as testing data;
	\item Select by signature: we use 257 mappings inferred by method signature, and select randomly a varying number of them as the training data and test against the remaining mappings in the ground truth.
\end{itemize}

The process repeats for using different folds as the training data for both Api2Api and we take the average accuracy are some observations from the results:

\begin{itemize}[leftmargin=*]
	\item Using the same number of seeds, either using the seeds from Random, K-fold or Signature-based, we get significantly better results than Api2Api for every setting.
	\item Our approach only needs 100 signature-based seeds to get a comparable result (43\% for Top-1) or better results (Top-5 and Top-10) with Api2Api that uses 688 manually crafted seeds.
	\item When using all of the 257 signature-based seeds, our approach gets significantly better results than Api2Api: top-1 improves 19\%, top-5 improves 34\%, and top-10 improves 27\%. %\yu{can we tell statistic significance? }
\end{itemize}

\paragraph{Compare to StaMiner and DeepAM}

We follow the details described in StaMiner and DeepAM to measure how well SAR performs in mining API mappings for Class API and Method API. In Java, an API element, by definition, can be a class, a method or a field in the class; and it must belong to a package (or the namespace in case of C\#). As such, the goal in this task is to measure the performance the Class and Method API mapping task one by one for each API of each package, i.e., to see which package has the best performance for API mappings, so-called 1-to-1 mappings. For the method API mapping, we use the 860 method ground truth mapping described in Section~\ref{sec:dataset} for evaluation. For the class API mapping, we use the 430 class ground truth mapping described in Section~\ref{sec:dataset} for evaluation. We follow the details described in DeepAM to choose only the APIs under the packages as shown in Table \ref{tab:1_1_mapping}, column 'Package', so that the total number of method API mapping left is 289 (remaining from 860 ground truth method API mappings), and the total number of class API Mapping 283 (remaining from 430 ground truth class API mappings).

Adapting SAR for class-level API mapping is relatively easy: one can remove the method part of a qualified API signature token so that only the package and class parts of the token are retained in the code sequences. Then code embedding for the API sequence can be derived as the embedding of the class-level API, along with other keywords from the ASTs. We do this for both languages. To select mapping seeds by API signatures, we first infer the mappings from signatures at the class level, then follow a similar domain adaptation process from APIs at the method-level. %The testing phase is the same as for the method API mapping.
% \n{revise wording}

One could not run StaMiner and DeepAM directly because they require parallel data (aligned function body for StaMiner, and aligned code and text description for DeepAM) for training. Therefore, we had to compare to them by extracting the reported performance numbers from their papers. This is also how DeepAM compared itself to StaMiner. 

We use the F-score as the performance metric to measure accuracy in this evaluation. It is defined as $F = (2PR)/(P + R)$, where Precision $P = TP/(TP + FP)$ and Recall $R = TP/(TP + FN)$. TP refers to the  number of true positives, which is the number of API mappings that are in both result datasets and the groundatasets; TN refers to the number of true negatives, which is the number of API mappings that are neither in the returned results nor in the ground truth datasets; FP refers to the  number of false positives which represents the number of result mappings that are not in the ground truth set; FN refers to the number of false negatives, which represents the number of mappings in the ground truth set but not in the results. 

Table~\ref{tab:1_1_mapping} shows the comparison results of our mined API mappings with StaMiner (Sta) and DeepAM (DeepA). Columns ``Class Mapping" and ``Method Mapping" list results of comparing API classes and methods, respectively. As one can see for the F-score, our approach has better results than those of DeepAM and StaMiner at the level of both classes and methods with much fewer seeds, while DeepAM needs to use millions of similar API sequence descriptions and StaMiner needs to use ten of thousands of pairs of parallel data.
%We do this with a ground truth dataset of manually written API mappings in Java2CSharp project for those JDK packages shown in the Table \ref{tab:1_1_mapping}.

%Table~\ref{tab:1_1_mapping} shows the comparison results of our mined API mappings with StaMiner (Sta) and DeepAM (DeepA). Columns "Class Mapping" and "Method Mapping" list results of comparing API classes and method, respectively". As one can see, the F-score at both class-level mapping and method-level mapping, SAR has better results than DeepAM and StaMiner.

\paragraph{Newly found API mappings}
More interestingly, we found a lot more new API mappings than other studies in our actual code corpora.
For each of the API in Java, we query the top-10 nearest neighbors in C\# and manually verify the mappings.
We enforce the threshold = 0.7 as mentioned in Section~\ref{sec:refinement_study} for this task.
%, for the score that is smaller than the threshold, the mapping is skipped, thus a lot of (likely) incorrect mappings are ignored, this reduces the human effort to verify the correctness of the mappings. 
We found 420 new SDK API mappings that can complement the tool Java2CSharp. Comparing to MAM (25 new mappings), StaMiner (125 new mappings), Api2Api (52 new mappings), we found a sufficiently larger number of mappings and our newly found APIs also overlap with the APIs in these baselines. 
%thus expecting that when applying such new APIs into Java2CSharp, the tool can reduce the error rate and increase the correctness when performing language migration. 
In~Table \ref{tab:newly_found}, we show some interesting examples of such newly found API mappings whose name do not match exactly using traditional approaches.
Our list of newly found Java/C\# APIs mappings can be accessed at this anonymous Github repository: \footnote{\url{https://github.com/djxvii/fse2019/blob/master/new_found/new_found_apis.csv}}.

\begin{table}
	\centering
	\caption{Examples of newly found APIs in Java and C\#}
	\label{tab:newly_found}
	\fontsize{5.5}{7}\selectfont 
	\begin{tabular}{|c|c|}\hline
		Java                 & C\# \\\hline\hline\hline
		java.io.DataInputStream.readInt & System.Io.BinaryReader.ReadUInt16  \\\hline
		java.awt.Graphics2d.fillRect & System.Drawing.Graphics.FillRectangle \\\hline
		javax.swing.Text.JtextComponent.setCaretPosition & System.Windows.Controls.RichTextbox.CaretPosition\\\hline
		java.lang.Byte.parseByte & System.sbyte.Parse \\\hline
		java.lang.Double.longBitsToDouble & System.BitConverter.Int64BitsToDouble \\\hline
		java.net.Datagramsocket.isConnected & System.Net.Sockets.Socket.Connect\\\hline
		java.awt.geom.AffineTransform.inverseTransform & System.Drawing.Drawing2d.Graphicspath.Transform \\\hline
		java.io.DataInputStream.readDouble & System.Io.BinaryReader.ReadDouble \\\hline
		java.net.Serversocket.accept& System.Net.Sockets.Socket.AcceptAsync
		\\\hline
	\end{tabular}
\end{table}

%the higher the threshold that is using, the higher the Precision. This can be useful in the cases that we focus on achieving higher accuracy at the expense of coverage. 

\subsubsection{RQ2. Effect of Different Refinement Approaches}
\label{sec:refinement_study}

\begin{table}
	\centering
	\fontsize{7.5}{10}\selectfont 
	\caption{Accuracy using various similarity thresholds}
	\label{tab:threshol_confidence}
	\begin{tabular}{|c|c|c|c|c|}
		\hline
		\multirow{2}{*}{Threshold} & \multicolumn{2}{c|}{Coverage} & \multicolumn{2}{c|}{Accuracy} \\ \cline{2-5} 
		& Top-1         & Top-5         & Top-1         & Top-5         \\\hline\hline\hline
		0.6                        & 0.76          & 0.94          & 0.42          & 0.59            \\ \hline
		0.7                        & 0.34          & 0.56          & 0.51          & 0.63            \\ \hline
		0.8                        & 0.10          & 0.19          & 0.65          & 0.80            \\ \hline
		0.9                        & 0.04          & 0.09          & 0.73          & 0.89            \\ \hline
	\end{tabular}
\end{table}

\paragraph{Effects of Cosine Similarity Threshold}
In this section, we measure the effect of different ways to combine the seeds for the refinement step. We want to measure the effects of cosine similarity threshold in order to choose a good one for the second heuristic in the refinement step. Since threshold is a part of the refinement, the domain adaptation step only comprises of two steps: Seeding and Adversarial Learning. Once the threshold is found, we use it for the Refinement in the other experiments.

Then we produce the mapping for each source query in the 860 ground truth mappings. For each mapping produce, we obtain the cosine similarity between the query and the result mapping. We choose a threshold to filter out the mapping that has the cosine similarity lower than the threshold, then we measure the accuracy of the left mappings.

In Table~\ref{tab:threshol_confidence}, the column "Coverage" means the percentage of ground truth APIs that have mappings in the candidate selection results when choosing a specific cosine similarity threshold.
The column ``Accuracy`` means the top-$k$ accuracy in identifying the mapping given a cosine similarity threshold as a condition to identify.
The results show that our approach in these experiments has higher mapping accuracy, but lower coverage with respect to the ground truth set when the similarity threshold increases. It is, therefore, a trade-off to have higher accuracy in the expense of coverage. For the other experiments that involve the cosine similarity threshold in the refinement, we choose 0.7 as the threshold as this number is balanced between the coverage and the accuracy.

\paragraph{Effects of Different Combinations of Refinement Heuristics}

Obtained 0.7 as a good threshold to identify correct mappings, we use this number for the "Cosine Similarity Threshold " heuristic in the Refinement step. What we measure is the impact of the two refinement heuristics on the performance, either using only one of them or combine them together. The domain adaptation also comprises of only Seeding and Adversarial Learning. After Adversarial Learning, we use different combinations of refinement heuristics to measure the effect of each heuristic. We use the 860 ground truth mappings from Java2CSsharp as the test set.

The results in Table~\ref{tab:refinement} show that taking the Intersection between the Top-K Frequency and the Cosine Threshold heuristic results in the best performance. This implies that the Cosine Threshold has an effect to filter out poor Top-K Frequency synthetic seeds, thus making the refinement better in overall.

\begin{table}
	\centering
	\fontsize{7.5}{10}\selectfont
	\caption{Different ways to combine refinement heuristics}
	\label{tab:refinement}
	\begin{tabular}{|c|c|c|c|}\hline
		Refine Method & Top-1 & Top-5 & Top-10 \\\hline\hline\hline
		
		Top-K & 0.53 & 0.70 & 0.78 \\ \hline
		
		Cosine & 0.24 & 0.29 & 0.36 \\ \hline
		
		Union Top-K + Cosine & 0.36 & 0.42 & 0.50 \\ \hline
		
		Intersection Top-K + Cosine  & 0.54 & 0.75 & 0.82 \\ \hline

	\end{tabular}
\end{table}

\subsubsection{RQ3: Effect of Overlapping Seeds}
\label{sec:overlapping_seeds}
As mentioned in Section~\ref{sec:implementation}, we found 257 seeds from the signature-based matching heuristic as the training data; however, 83 of them overlap with the ground truth set. These 83 overlapping seeds could affect the final performance even though these 257 seeds are part of our end-to-end approach. To analyze the effect of these overlapping seeds, we also remove them from the training data, using the remaining 174 seeds to evaluate. Comparing the results to those using all 257 seeds as the training data, Table~\ref{tab:effect_of_overlapping} shows that accuracy suffers a few percents, but the overall score is still fairly good.

\begin{table}
	\centering
	\fontsize{8}{10}\selectfont 
	\caption{Effect of overlapping seeds}
	\label{tab:effect_of_overlapping}
	\begin{tabular}{|c|c|c|c|c|}\hline
		Baselines &Seeds& Top-1 & Top-5 & Top-10 \\\hline\hline\hline
		Ours: Removed overlapping seeds &174 & 0.51 & 0.69 & 0.76\\\hline    
		Ours: All seeds                 &257 & 0.54 & 0.75 & 0.82\\\hline
	\end{tabular}
\end{table}

\subsubsection{RQ4: Effect of Each Component}
We performed an ablation study of domain adaptation to measure the performance of individual components as well as their combinations (Table~\ref{tab:ablation_study}). %We measure the performance of SAR with all of the combinations of any of the two components. We also measure performance when each component stands alone.
Note that for the Refinement component, since Section~\ref{sec:refinement_study} shows that using the intersection of Top-K and cosine threshold leads to better results than union, we refer Refinement to those of ``Intersection of Top-K and Cosine" performance.

Here are some observations from the results:

\begin{itemize}[leftmargin=*]
	\item Seeding is the most important step for the domain adaptation to works well, e.g even with a small set of seeds (25), which is a very small knowledge, it sets up a basis for the adversarial learning to improve the performance significantly.
	\item Adversarial Learning is essential in improving the performance, e.g comparing Seeding with Seeding-Adv, the top-1 accuracy is improved by 19\% on average. 
	\item Without the Adversarial Learning, either the Seeding or the Refinement step alone, or combine these two together does not achieve good performance.
	\item Refinement alone does not achieve any good result because the initial input matrix was completely random that cannot be refined to anything better; 
	%\n{need to revise this part}
	\item Using the Adversarial Learning alone achieves some reasonable results, e.g top-1  = 24\%, top-10  = 35\%. Further with the Refinement step, top-1 improves to 29\%, top-10 becomes 40\%. These can be seen as the results of unsupervised domain adaptation without any initial seeds. 
\end{itemize}

\begin{table}
	\centering
	\fontsize{7}{8}\selectfont 
	\caption{Ablation Study -- effects of each component}
	\label{tab:ablation_study}
	\begin{tabular}{|c|c|c|c|c|}\hline
		Baselines &Seeds& Top-1 & Top-5 & Top-10 \\\hline\hline\hline        
		\multirow{2}{*}{Seeding+Adv}  
		&25 & 0.30 & 0.39 & 0.45 \\ \cline{2-5}
		&50 & 0.32 & 0.40 & 0.54 \\ \cline{2-5}
		&100& 0.40 & 0.53 & 0.71 \\ \cline{2-5}
		&257& 0.51& 0.73 &0.82 \\ \hline
		
		\multirow{2}{*}{Seeding+Refine}  
		&25 & 0.13 & 0.15 & 0.20 \\ \cline{2-5}
		&50 & 0.20 & 0.23 & 0.29 \\ \cline{2-5}
		&100& 0.25 & 0.29 & 0.43 \\ \cline{2-5}
		&257& 0.33 & 0.40 & 0.53 \\ \hline
		
		\multirow{2}{*}{Seeding}  
		&25 & 0.12 & 0.15 & 0.18 \\ \cline{2-5}
		&50 & 0.20 & 0.19 & 0.28 \\ \cline{2-5}
		&100& 0.23 & 0.27 & 0.36 \\ \cline{2-5}
		&257& 0.32 & 0.39 & 0.42\\ \hline
		
		Refine &- & 0.01 & 0.01 & 0.01\\\hline              
		Adv+Refine &- & 0.29 & 0.34 & 0.40\\\hline    
		Adv       &- & 0.25 & 0.30 & 0.35\\\hline    
		
	\end{tabular}
\end{table}

%\begin{table}
%    \centering
%    \caption{Results of API Sequence translation retrieval}
%    \label{tab:api_sequence_retrieval}
%    \begin{tabular}{|c|l|c|c|c|}\hline
%        Project     & Baselines        & P@1 & P@5 & P@10 \\\hline\hline\hline
%        Antlr     & Api2Api            & 0.12 & 0.18 & 0.26\\\hline
%        & Gan-Synth Refine    & 0.12 & 0.18 & 0.26\\\hline
%        db4o     & Api2Api            & 0.12 & 0.18 & 0.26\\\hline
%        & Gan-Synth Refine    & 0.12 & 0.18 & 0.26\\\hline
%        Fpml     & Api2Api            & 0.12 & 0.18 & 0.26\\\hline
%        & Gan-Synth Refine    & 0.12 & 0.18 & 0.26\\\hline
%        Itext     & Api2Api            & 0.12 & 0.18 & 0.26\\\hline
%        & Gan-Synth Refine    & 0.12 & 0.18 & 0.26\\\hline
%        JGit     & Api2Api            & 0.12 & 0.18 & 0.26\\\hline
%        & Gan-Synth Refine    & 0.12 & 0.18 & 0.26\\\hline
%        jts     & Api2Api            & 0.12 & 0.18 & 0.26\\\hline
%        & Gan-Synth Refine    & 0.12 & 0.18 & 0.26\\\hline
%        Lucene     & Api2Api            & 0.12 & 0.18 & 0.26\\\hline
%        & Gan-Synth Refine    & 0.12 & 0.18 & 0.26\\\hline
%        Neodatis& Api2Api            & 0.12 & 0.18 & 0.26\\\hline
%        & Gan-Synth Refine    & 0.12 & 0.18 & 0.26\\\hline
%        POI     & Api2Api            & 0.12 & 0.18 & 0.26\\\hline
%        & Gan-Synth Refine    & 0.12 & 0.18 & 0.26\\\hline
%        
%        
%    \end{tabular}
%    
%    
%\end{table}

\subsection{Explainability Analysis of the Results}
We performed various explainability analyses of our model in varying configurations to obtain some insights about our method. From the results, we show that our approach performs significantly better than Api2Api in every perspective. An interesting question one may ask is "\textit{why does this approach perform better than Api2Api}?". Although theoretically, Adversarial Learning maximizes the similarity between two distributions, it is still useful to explain this phenomenon using analysis of the results.

\subsubsection{Effect of Refinement on Frequent vs Rare tokens}
We note that the frequency of an API token could affect the quality of the mapping result, i.e more frequent tokens could affect performance more than the less frequent ones. With this assumption, the Refinement of the mapping matrix tries to improve the mapping by using frequent tokens as the anchor. To measure the effect of the refinement on the frequent tokens and rare tokens, we ranked the 860 ground truth mappings in Java2CSharp by the frequency of the source APIs, i.e. the Java JDK APIs. Then we use our model to produce the mapping results against the top 10\%, which is a subset of frequent tokens; and bottom 10\%, which is a subset of rare tokens. To ensure a fair comparison, we use the 174 non-overlapping seeds in Section~\ref{sec:overlapping_seeds} to train the domain adaptation procedure.

\begin{table}
	\centering
	\fontsize{7}{8}\selectfont 
	\caption{Effect of Refinement on Frequent vs. Rare Tokens}
	\label{tab:frequent_tokens}
	\begin{tabular}{|c|c|c|c|c|c|}
		\hline
		\multirow{2}{*}{Baselines}      & \multirow{2}{*}{\% Ground truth} & \multirow{2}{*}{Eval size} & \multicolumn{3}{c|}{Accuracy} \\ \cline{4-6} 
		&                                  &                            & Top-1    & Top-5   & Top-10   \\\hline\hline\hline
		\multirow{2}{*}{With Refine}    
		& Top 10\%                   & 86                         & 0.65     & 0.78    & 0.85     \\ \cline{2-6} 
		& Bottom 10\%               & 86                         & 0.32     & 0.35    & 0.47     \\ \hline
		\multirow{2}{*}{Without Refine} & Top 10\%                         & 86                         & 0.54     & 0.65    & 0.72     \\ \cline{2-6} 
		& Bottom 10\%                      & 86                         & 0.30     & 0.34    & 0.45     \\ \hline
	\end{tabular}
\end{table}

The results in Table \ref{tab:frequent_tokens} show the following observations:
\begin{itemize}[leftmargin=*]
	\item Mapping accuracy decreases while increasing top-k frequent tokens in the evaluation set, in either setting. This implies that token frequency does affect on the mapping result;
	\item The refinement step can improve the result of both the frequent tokens and rare tokens, although the impact is bigger on frequent tokens, e.g., improved by 10\% for top-10\% , and only 2\% for bottom-10\%.
\end{itemize}

\subsubsection{Retrieved Results Comparison}

\begin{table*}
	\caption{Retrieved API mapping results from sample queries produced by SAR and Api2Api.}
	\fontsize{7}{8.5}\selectfont 
	\label{tab:retrieval-example}
	
	\begin{tabular}{|c|c|}
		\hline
		\textbf{SAR}                                               & \textbf{Api2Api}                                              \\ \hline
		\midrule
		\multicolumn{2}{|c|}{\textbf{(1) java.util.Collection.add}}                                                                    \\ \hline
		\textbf{System.Collections.Objectmodel.Collection.Add}     & \textbf{System.Collections.Generic.List.Add}                  \\ \hline
		\textbf{System.Collections.Generic.List.Add}               & System.Collections.Generic.List.Get                           \\ \hline
		System.Collections.ObjectModel.Collection.Clear            & System.Collections.Generic.List.Remove                        \\ \hline
		System.Collections.Generic.List.Contains                   & \textbf{System.Collections.Objectmodel.Collection.Add}        \\ \hline
		\textbf{System.Collections.Generic.Dictionary.Add}         & System.Collections.IDictionary.GetEnumerator                  \\ \hline
		\midrule
		\multicolumn{2}{|c|}{\textbf{(2) javax.swing.Text.JtextComponent.setCaretPosition}}                                            \\ \hline
		System.Windows.Controls.RichTextBox.Clip                   & System.Drawing.Image.GetframeCount                            \\ \hline
		System.Web.Ui.Webcontrols.DataGrid.PageSize                & System.Media.SoundPlayer.PlaySync                             \\ \hline
		\textbf{System.Windows.Controls.RichTextBox.CaretPosition} & System.Web.Ui.Webcontrols.Calendar.WeekendDayStyle            \\ \hline
		System.Windows.Forms.ContextMenuStrip.SuspendLayout        & System.Configuration.Xmlutil.StrictSkipToNextElement          \\ \hline
		System.Windows.Controls.RichTextBox.CaretBrush             & System.Media.SoundPlayer.PlayLooping                          \\ \hline
		\midrule
		\multicolumn{2}{|c|}{\textbf{(3) java.io.File.exists}}                                                                         \\ \hline
		\textbf{System.Io.File.Exists}                             & \textbf{System.Io.File.Exists}                                \\ \hline
		System.Io.File.AppendText                                  & System.Web.Errorformatter.ResolveHttpFileName                 \\ \hline
		System.Io.File.Delete                                      & System.Io.File.OpenRead                                       \\ \hline
		System.Io.Fileinfo.LastWriteTime                           & System.Io.Compression.Zipfile.OpenRead                        \\ \hline
		System.Io.File.GetAttributes                               & System.Io.Compression.ZipFile.ExtractToDirectory            \\ \hline
		\midrule
		\multicolumn{2}{|c|}{\textbf{(4) java.util.concurrent.atomic.AtomicInteger.getAndDecrement}}                                   \\ \hline
		System.Threading.Interlocked.Decrement                     & System.Directoryservices.SearchResultCollection.GetEnumerator \\ \hline
		System.Threading.ReaderWriterLockSlim.EnterWriteLock       & System.Directoryservices.SearchResultCollection.Dispose       \\ \hline
		System.Threading.Interlocked.Increment                     & System.Runtime.Serialization.ObjectIdGenerator.HasId          \\ \hline
		System.Threading.EventWaitHandle.OpenExisting              & System.Collections.Generic.Queue.CopyTo                       \\ \hline
	\end{tabular}
\end{table*}
\begin{sloppypar}
	To evaluate our approach qualitatively, we retrieved C\# API methods from sample queries in Java SDK. Table~\ref{tab:retrieval-example} shows the resulting top-5 C\# APIs for four queries: $java.util.Collection.add$, $java.io.File.exists$ $javax.swing.Text.JTextComponent.setCaretPosition$, and $java.util.concurrent.atomic.AtomicInteger.getAndDecrement$. 
	They are ordered by increasing difficulty in finding a mapping. %The first query has multiple mappings, the second query has only one exact mapping, the third query also has one exact mapping, but the name of the mapping is not exactly the same as the query, and the fourth query has no corresponding mapping at all.
	
	For the first query, we can see that both Api2Api and our approach can successfully select the correct top-1 mapping, the other results are also related. This case can be considered as easy for both approaches to performing well.  
	
	For the second query, both approaches can achieve a good exact mapping, but for the other results, our approach can generalize all of the results under the `$System.IO.File$' class, while there are some less related results in the top-5 produced by Api2Api, e.g `$System.Web.ErrorFormatter.ResolveHttpFileName$'.
	
	The third query token ranks the 11,204th in the embedding table\footnote{The order of the token embedding provided by word2vec is proportional to the frequency of the token~\cite{mikolov2013distributed}}. As discussed earlier, embedding quality of rare tokens is not as good as those of frequent tokens. Therefore, it is more difficult to find an exact mapping for such a query. Even so, our approach can still rank a correct mapping at the third place (`$System.Windows.Controls.RichTextBox.CaretPosition$'), while Api2Api produce totally unrelated results.  
	
	For the last query, even though there has no mapping in C\# by the ground truth, the retrieved results are still reasonably close. The query, in this case, is an API for an atomic operation, which is related to thread handling. Our approach can generalize the result mappings to the `$System.Threading$' APIs in C\#, while the results from Api2Api are totally unrelated. 
	
	This experiment shows that Adversarial Learning can maximize the similarity between the two distributions so that similar APIs are clustered together. 
\end{sloppypar}    

\subsubsection{API Clustering Ability}

We perform an additional analysis at the package level to show the ability of Adversarial Learning to cluster similar APIs under the same package, e.g., `$java.io$' APIs in Java should be close to `$System.IO$' APIs in C\#. To define the ground truth of the aligned packages, we refer to the 860 ground truth mappings from Java2CSharp datasets: when more than 50\% of APIs in a package in one language have the corresponding of APIs in another language, we say that these two packages are aligned. For example, we say that the package `java.io' and `$System.IO$' are aligned because 83\% of `$java.io$' APIs in Java is aligned with some `$System.IO$' APIs in C\#. In total, we can derive 5 pairs of aligned packages: (1) $java.io$ -- $System.IO$, (2) $java.math$ -- $System.Math$, (3) $java.net$ -- $System.Net$, (4) java.sql -- System.Data.SqlClient, and (5) $java.util$ -- $System.Collections.Generic$.

\begin{table}
	\caption{Average Cosine Similarity Comparison}
	\label{tab:cluster_distance}
	\begin{tabular}{|c|l|l|l|l|l|}
		\hline
		\multirow{2}{*}{Baselines} & \multicolumn{5}{c|}{Average Score Per Package}                                                                                                              \\ \cline{2-6} 
		& \multicolumn{1}{c|}{java.io} & \multicolumn{1}{c|}{java.math} & \multicolumn{1}{c|}{java.net} & \multicolumn{1}{c|}{java.sql} & \multicolumn{1}{c|}{java.util} \\ \hline
		Api2Api                    & 0.23                         & 0.66                           & 0.27                          & 0.31                          & 0.17                           \\ \hline
		SAR               & 0.39                         & 0.73                           & 0.58                          & 0.48                          & 0.41                           \\ \hline
	\end{tabular}
\end{table}

Using these data, we first compute the pairwise cosine similarity scores of all pairs of APIs under each pair aligned packages, then take the average of the scores. We do this for both Api2Api and our approach. Table~\ref{tab:cluster_distance} shows that the average scores produced by our approach are significantly better than those of Api2Api, which implies that our approach has the ability to cluster the similar group of APIs together.

%\subsection{Discussions}

% Please add the following required packages to your document preamble:
% \usepackage{multirow}

%The ultimate goal of our paper is to proposed a zero-knowledge domain adaptation approach for any pairs of languages, yet in this paper, we only perform the experiments on Java and C\# as we found it not easy to find a good and large enough evaluation dataset for the other pairs of languages. As a result, we need to spend some human efforts to find and manually verify the mappings to get a good evaluation dataset. We leave this task in the future. 
%
%The synthetic seeds we inferred may not be that many for other pairs of languages. The reasons can be due to the difference in coding styles of the language, e.g $binary\_search$ and $binarySearch$, which is not an exact match, or can be due to some more additional information in the method, e.g $readInt$ and $readInt16$ . In the future, we may relax the matching condition by involving a sequence matching algorithm, such as edit distance. With this, we hope to find more seeds and the process is more flexible for multiple languages. 
%
%We only consider the sequence that contains at least 5 API tokens to estimate the OOV. In reality, such case may not happen frequently as not all the time there are more than 5 function calls in a context. In the future, we will consider more contexts, such as Call Graph, Control Flow Dependency, or even the internal implementation of the OOV as the context, so that we can have enough context to estimate the OOV.

%% file: threats.tex
%!TEX root=sample-sigconf.tex
The goal of domain adaptation is to use as little knowledge as possible for any pair of languages. However, we only perform the experiments on Java and C\# in this paper because it is not easy to find a good and large enough evaluation dataset for other pairs of languages. We leave this task in the future. 

%\lx{the "risk" of dissimilar distributions of vector spaces.}
While unsupervised adversarial learning method does not require any seed as parallel data, there is a risk that the distributions of vectors (embeddings) in the two spaces are not so similar. Through our experiments, it is confirmed that the performance could be improved further by initializing the unsupervised adversarial learning method with a small set of seeds taken from the seed-based domain adaptation, and by generating the rest of API mappings. 

One limitation of our approach is that we can only generate single API mapping instead of an API sequence mapping. Both Api2Api and ours share such a limitation. In Api2Api, they use the new mappings mined from the tool as the input for an external machine translation tool, Phrasal~\cite{Phrasal}, to generate the mapping for API sequences. %By a design similar to Api2Api \n{can we use a better word than 'similar' to avoid the impression that we are similar o Api2Api}, 
In the future, we can also feed the newly found mapping APIs from our tool to Phrasal as inputs.
%To control the quality of the API sequence that such a tool provides, we plan to develop a self-contained technique that does not depend on third-party tool to use our output. 

%Our goal in the future is to develop a new algorithm that requires no seed at all for API migration.
%\lx{recall what we discussed, this future work doesn't make sense. Revise.} 
%In NLP, there is various work for single word translation under the unsupervised setting (almost or no parallel data), which served as the cornerstone to achieving the unsupervised machine translation task. Our goal is about the same, we aim to develop a system that can automate the API migration process without any human intervention.

We mainly use a simplified top-k accuracy metric to measure our performance against the Api2Api. In real-world use cases, other information retrieval based metrics, such as MAP and MRR, may have less bias in evaluating the list of API mappings. We leave this for the future.

%% file: conclusion.tex
%!TEX root=sample-sigconf.tex
We have proposed a domain adaptation approach, named SAR, to automatically transform and align the vector spaces used to represent two different languages and APIs used therein.
We adapt code embedding and adversarial learning techniques with a seeding and refinement method to implement our approach.
The approach can identify API mappings across different programming languages.
Our evaluation shows that the mappings between Java and C\# APIs identified by our approach can be more accurate than other approaches with just 257 mapping seeds that can be easily identified by an automatic, simple signature-based heuristic, and it helps to identify hundreds of more API mappings between Java and C\# SDKs.
% our prototype can achieve much better performance than the state-of-the-art technique. 

Domain adaptation methods are useful for other software engineering tasks that involve two different domains targeted by transferred learning~\cite{Nam2018,Nam2013,Yan2017}, such as cross-language program classification, code summarization, cross-language/project bug prediction. These tasks may benefit from the proposed approach when little curated data is available. Other SE tasks that are challenging due to lack of data, such as the out-of-vocabulary (OOV) problem~\cite{Cvitkovic2018,Allamanis2018,hellendoorn2017fse} for learning and modeling fast-evolving software code, may also benefit from our domain adaptation approach, 
because the embeddings of OOV words may be approximated on-the-fly by adapting the known embeddings of their contextual or similar words in different languages. 
In the future, we will explore these variants of applications.